\documentclass[manuscript,screen]{acmart}


\AtBeginDocument{%
  }

\setcopyright{acmlicensed}
\copyrightyear{2018}
\acmYear{2018}
\acmDOI{XXXXXXX.XXXXXXX}
\acmConference[AI Letters]{}
\acmISBN{978-1-4503-XXXX-X/2018/06}

\begin{document}

\title{From Performance to Purpose: A Sociotechnical Taxonomy for Evaluating Large Language Model Utility}

\author{Gavin Levinson}
\email{gavinlev@umich.edu}
\affiliation{%
  \institution{University of Michigan}
  \city{Ann Arbor}
  \state{Michigan}
  \country{USA}
}
\author{Keith Feldman}
\email{keithfel@med.umich.edu}
\orcid{0000-0001-6935-5844}
\affiliation{%
  \institution{University of Michigan}
  \city{Ann Arbor}
  \state{Michigan}
  \country{USA}
}

\begin{abstract}

As large language models (LLMs) continue to improve at completing discrete tasks, they are being integrated into increasingly complex and diverse real-world systems. However, task-level success alone does not establish a model’s fit for use in practice. In applied, high-stakes settings, LLM effectiveness is driven by a wider array of sociotechnical determinants that extend beyond conventional performance measures. Although a growing set of metrics capture many of these considerations, they are rarely organized in a way that supports consistent evaluation, leaving no unified taxonomy for assessing and comparing LLM utility across use cases. To address this gap, we introduce the Language Model Utility Taxonomy (LUX), a comprehensive framework that structures utility evaluation across four domains: performance, interaction, operations, and governance. Within each domain, LUX is organized hierarchically into thematically aligned dimensions and components, each grounded in metrics that enable quantitative comparison and alignment of model selection with intended use. In addition, an external dynamic web tool is provided to support exploration of the framework by connecting each component to a repository of relevant metrics (factors) for applied evaluation. 

\end{abstract}



\maketitle

\section{Introduction}

Large language models (LLMs) have emerged as a revolutionary technology, shaping decision making, automating tasks, and redefining how individuals and organizations interact with information~\cite{xie2025social}. Driven by rapid advancements in natural language processing techniques, availability of massive datasets, and enormous investments in compute resources, there has been a proliferation of LLMs, each presenting an opportunity to optimize for specific knowledge bases and applications~\cite{singhcomparison}. While this array of specialized models offers significant promise for translating technical advancements to real-world utility, it also presents a considerable operational challenge in determining which model may be optimal for a given use case~\cite{zhangwe,apollinaire2024choose}.


In making such determinations, efforts have largely focused on the performance of LLMs. Giving rise to initiatives such as Stanford’s Holistic Evaluation of Language Models (HELM)~\cite{liang2022holistic}, the Massive Multitask Language Understanding (MMLU) suite~\cite{hendrycks2020measuring}, or several other systematic approaches comparing model accuracy, completeness, and reliability across a sundry of analytical tasks. Yet this approach conflates technical performance with model operation in real-world contexts~\cite{raji2021ai}. In practice, utility of LLMs is driven by a wide range of technical, social, and policy factors~\cite{kudina2024sociotechnical,xu2024intelligent}.


This inconsistency between metrics and usage has brought forth a host of sociotechnical considerations from both the research and practitioner communities, spanning industry leaders advocating for economic consideration in the training and execution of models~\cite{kandpal2025position}, to privacy experts demanding processes to assess safety~\cite{ajuzieogueconomic,oseni2021security}. These appeals mark a fundamental shift in LLM benchmarking, indicating a clear need to move beyond simply the correctness of operations performed. Rather, assessment of LLMs must reflect a broader set of \textit{utility} dimensions. In much the same way dimensions of data quality are deeply tied to the efficacy of downstream analytics~\cite{declerck2024frameworks,schwabe2024metric}, the potentiality of LLMs for a given task can be organized into a set of domains encompassing both functional and operational elements of a models’ deployment.  

To date, technical benchmarks have flourished due to their ability to standardize model comparison; as LLMs are deployed in larger and more complex systems, however, assessing utility will require not just a wider array of factors, but also an organization by which to consider them~\cite{eriksson2025can}. Accordingly, this work introduces the Language Model Utility Taxonomy (LUX), a comprehensive framework for evaluating LLM utility ranging from granular technical elements of the model architecture to broad, high-level considerations of workflow integration.

This contribution advances the consideration of LLM utility beyond a traditional measure of technical performance to a framework that synthesizes expanded utility dimensions with the metrics used to quantify them. In doing so, it helps practitioners and organizations align model selection with specific needs by providing a structured understanding of LLM capabilities. 


\section{Framework Overview}
The LUX framework spans four overarching domains, each defining a key theme of model utility: performance, interaction, operations, and governance. Each domain is organized into a hierarchical structure, broken down into dimensions then components that translate high-level constructs into tangible criteria. Ultimately, components are defined by specific metrics, which are designed to offer quantitative evaluation of a model. To understand the state of the field, a review of current metrics for each component can be found in Appendix B. This hierarchical structure enables utility to be examined at multiple levels, allowing comparisons both within and across domains.

At the foundation of LLM operation, and driving properties of the four domains, is the core element of model training. Training is made up of two pillars: data and model architecture~\cite{zhuang2017challenges}. Data determines the depth and breadth of information available to the model. These properties dictate what the model learns, the biases it inherits, and the contexts it can generalize across \cite{zhou2025survey}. Contemporary models increasingly draw from live external sources rather than static, pre-trained datasets~\cite{singh2025agentic}. This development expands capabilities but requires safeguards to ensure retrieved information is trustworthy and aligned with safe deployment practices~\cite{yang2024crag}. Model architecture governs how information is represented and transformed into output by defining the internal mechanisms that enable generation~\cite{koehler2024large}. During training, these elements interact with data to produce the behaviors that shape how a LLM performs in applied settings~\cite{liu2025understanding}.

This article offers a description of LUX domains and dimensions, while a dynamic web tool\footnote{https://chart-lab.github.io/LUX-Framework/} has been created allowing users to explore the framework organization from a high-level down to a repository of relevant metrics for each component. 

\section{Domain}
\subsection{Performance}

A core principle of LLM utility remains its performance, broadly defined by the capacity of a model to accomplish tasks in a given context~\cite{arabzadeh2024assessing}. Yet, in contrast to models that yield discrete and verifiable outputs, evaluation of generative models is complex and must capture the nuanced interplay between open-ended queries and non-deterministic responses~\cite{song2025good}. As a result, performance must consider not only correctness, but also temporal, contextual, and behavioral factors that influence the quality of a model’s output~\cite{huang2025decmetrics}. 

\textbf{Task validity:} The cornerstone of traditional performance assessment, validity of LLM output, can be quantified across three discrete components. First, accuracy, which captures the factual correctness of information provided by a model. This determination spans both the objective precision of models’ response, as well as occurrences of hallucination~\cite{hu2024unveiling}. Second, the notion of completeness which considers how fully a response addresses all aspects of a query needed to meet the intended use~\cite{patil2025framework}. Third, is the timeliness of information provided by a model with respect to the query expectations~\cite{turkmen2025balancing}.

\textbf{Stability:} LLMs are inherently stochastic systems that can yield divergent responses, even when inputs are similar. Stability encapsulates a LLM’s ability to deliver consistent, reliable outputs across variations in inputs and interactions~\cite{shyr2025statistical}. The dimension consists of reproducibility, which captures the consistency of output when faced with identical prompts~\cite{tahmasebi2025fact}. When viewed at multiple time points, this component also captures the impact of architectures using feedback for dynamic updates to model weights. This idea is extended through a second component for robustness, which measures a model’s resilience to variability in semantically similar queries~\cite{chaudhary2024towards,fastowski2025confidence}. Finally, model confidence captures how well a LLM communicates uncertainty and resists user persuasion, which matters in an engagement-driven landscape where overly compliant models can propagate erroneous information~\cite{khanmohammadi2025calibrating,devic2025calibration, jain2025beyond}. 

\subsection{Interaction}
The interaction domain characterizes the intricate relationship between end-users, LLMs, and the systems in which these tools operate~\cite{schreiter2025evaluating}. Although input and output requirements vary across models and use-cases, the generative architecture underlying LLMs brings forth a need to quantify how information is processed and communicated~\cite{liu2024we,mu2024clarifygpt}.

\textbf{Workflow:} The fundamental form of interaction captures the logistics of how LLMs connect to data and users, mapping across two components. First, system integration, which takes guidance from platform engineering to quantify how LLMs connect with external systems~\cite{li2023beyond}. This spans both low-level technical integrations (e.g., API to stored data), and the broader set of systems a LLM can exchange information with, a central capability in emerging agentic AI~\cite{tupe2025ai}. Second, a component of interface, which defines how end-users engage with a model. Drawing from human factors evaluation, these measures offer guidance on user interaction and LLM operating environments~\cite{ambacher2024designing}.

\textbf{Presentation:} More than the technical interface itself, utility of generative models is largely influenced by \textit{how} information is conveyed to users. Tied closely to the modality of a response, the dimension of presentation takes direction from information design to assess clarity and comprehension of output~\cite{hu2025multimodal}. For text this may capture linguistic characteristics, while other modalities address a model’s  semantic and stylistic choices~\cite{chevi2025individual}. Further, this dimension includes measures of accessibility of created information, and alignment to output criteria and format expectations.
 
\textbf{Traceability:} Bridging operations and output, the dimension of traceability focuses on the ability of a LLM to establish a lineage of information created through generative methods. It serves as an essential determinant of trust for a system and is defined by elements of attribution and reasoning~\cite{barrak2025traceability}. In many cases, LLM outputs are only reliable to the extent that users can verify \textit{why} a model produced them. One component, leveraging scholarship in information retrieval, captures whether claims can be consistently linked to supporting source material. A second focuses on process visibility, which considers whether a model can surface a legible rational, as seen in chain-of-thought or other reasoning models~\cite{procko2025provenance}.

\subsection{Operations}
While other domains focus on behavioral properties, operations grounds utility in the practical constraints that shape implementation and maintenance. In doing so, it emphasizes how even a technically capable model can fail to deliver value if it cannot operate economically or meet time and scale requirements~\cite{gupta2025ai}. 

\textbf{Cost:} The economics of a model is a primary determinant of whether a system can be effectively adopted and scaled~\cite{davis2023local}. Contemporary LLMs increasingly incur large usage costs, which arise from subscription pricing and cost-per-token fees~\cite{shekhar2024towards}. Beyond direct usage costs, the underlying infrastructure required to run a model, including data, specialized hardware, energy, and support resources, can also be costly~\cite{cottier2024rising, luccioni2024power}. These factors are shaped by deployment choices such as hardware and geographic location, which influence monetary expenses, but also environmental and social externalities associated with AI computation~\cite{mulita2024societal}. As a result, cost reflects both immediate financial burden and long-term sustainability considerations, making it a central consideration of utility~\cite{pan2025cost}.

\textbf{Efficiency:} A LLM's capacity to deliver timely and resource-conscious outputs is vital for usage in fast-paced operational contexts. Efficiency is evaluated through components of speed, which considers how quickly a model produces responses~\cite{kang2025win}, and concurrency, which examines a models ability to handle multiple simultaneous users and tasks~\cite{yao2024scalellm}. These properties are not intrinsic to model architecture alone, but depend on multiple interacting factors that combine the complexity and scale of user requests with the location of model deployment~\cite{xiong2025high,adenekan2023optimizing}. An operationally efficient model must therefore sustain acceptable responsiveness and reliability as demand scales and environmental constraints vary.

\subsection{Governance}
Governance represents how a model manages its data, enforces constraints, and operates within institutional and regulatory boundaries. These concepts situate the more technical domains of performance and interaction within their operational context, evaluating not only a model’s functional capabilities, but the appropriateness and conditions of its application within its embedded environment~\cite{liu2025govbench}. A LLM that fails to adhere to guard rails erodes the foundation of safe integration, reducing trust \cite{bach2024systematic,holliday2016user} and undermining the credibility of both the technology and the organization employing it~\cite{akheel2025guardrails}.

\textbf{Policy Enforcement:} The extent to which a LLM aligns with standards governing its deployment underlies real-world operations. Specifically, a model must adapt to dynamic regulations and accountability structures that originate from multiple authorities globally~\cite{wang2025llm}. These guard rails function as constraints on a LLM's permissible scope of action; enforcing them ensures that generative behavior remains aligned with institutional intent, legal mandates, and safety standards~\cite{hassani2024enhancing}. Because these constraints are enforced through probabilistic mechanisms rather than learned rules, LLM systems remain vulnerable to jailbreaking~\cite{wei2023jailbroken}, and assessment of risk is a key component to implementation assessment. 

\textbf{Security:} The mechanisms with which a LLM safeguards the confidentiality of its data, interactions, and operations underscores its capacity to operate safely in real-world settings. Within this domain, privacy considers the treatment of data, ensuring that exchanged information remains contained within the appropriate boundaries~\cite{yao2024survey,wu2024new}. By contrast, statefulness examines how prior context is preserved and applied across interactions, balancing the benefits of sustained coherence with the risks of excessive retention~\cite{roy2024towards,biswasstateful}. 

\section{Conclusion}
At its core, the LUX framework marks a meaningful shift in our understanding of what it means to evaluate large language models. Recently, technical performance has dominated the metrics against which models are optimized and measured. This focus stemmed from the availability of objective and quantitative targets, which provided a clear path for improving model capability. Yet as LLMs are adopted across a growing range of domains, from scientific discovery to creative industries, it is clear that performance alone does not ensure success in complex, real-world environments. Addressing this gap requires us to realign the goals of evaluation, from outcomes to a broader paradigm of utility. By introducing a novel taxonomy of factors at the intersection of technical, institutional, and human domains, LUX illuminates model strengths and limitations through a more holistic lens.

Reflecting on evaluation more broadly, the importance of information and impact of bias permeates all domains and factors. Intertwined with user interactions, the information these models draw on, either learned from training or accessed during use, influences how they behave, and biases they may exhibit. Meaningful evaluation requires examining models in context. 

Ultimately, the framework takes the first steps toward a more adaptive approach to evaluation. As LLMs are dynamic systems, the measures used to assess their utility must also evolve. New applications, shifting user expectations, and changing institutional requirements demand that our evaluative criteria continue to expand. LUX is designed as a flexible framework, not a metric; enabling practitioners to incorporate emerging considerations and new sociotechnical factors, so that assessments remain rigorous and relevant as both models and their contexts change.

\newpage
\bibliographystyle{ACM-Reference-Format}
\bibliography{References}

\newpage
\appendix

\section{Appendix}

\subsection{LUX Figure}

\begin{figure}[ht]
  \centering
  \includegraphics[width=\linewidth]{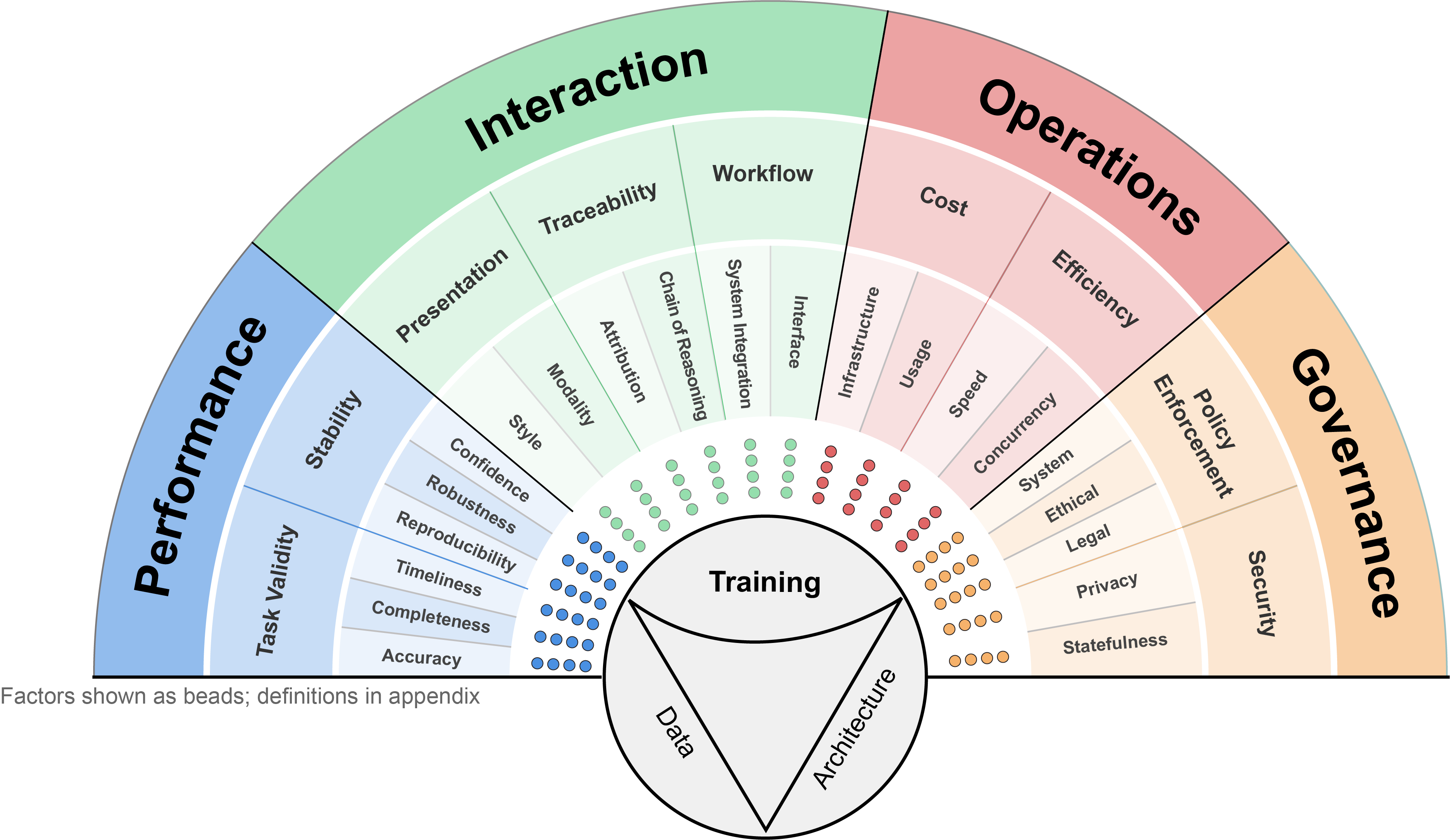}
  \caption{LUX, shown as a hierarchical framework, is organized from high-level domains (outer layer) inward to dimensions and then components (inner layer). Factors quantifying each component (represented as beads) are defined in the companion web tool.}
\end{figure}

\newpage
\section{Metrics}
This appendix describes and explores a representative body of contemporary metrics designed to quantify each component within the four domains of the LUX framework: performance, interaction, operations, and governance. By systematically cataloging these metrics, this section bridges the conceptual taxonomy developed in the main text with practical application, equipping researchers and practitioners with actionable criteria for model evaluation. For additional depth and ongoing updates, a more comprehensive set of metrics, along with descriptions of their measures and links to relevant articles/implementations, is available through the interactive online companion tool.

\subsection{Performance:} Performance benchmarking remains the most comprehensive and objective evaluation technique available to LLMs today. Drawing on a wide body of literature in natural language processing, validity is quantified through a combination of automated and human-centric metrics. Traditional scoring (F1~\cite{lipton2014thresholding}, BLEU~\cite{reiter1bleu}, etc.) is often compared against curated reference datasets, while metrics such as context recall \cite{zhang2019bertscore} and faithfulness \cite{fabbri2022qafacteval} quantify elements of completeness based on underlying source material. LLM-as-a-judge assessments are also evolving to assess open-ended queries and measure behavior such as the timeliness of information and hallucinations \cite{kasai2023realtime}. Stability has largely been operationalized by metrics using repeated queries and parameter tuning to assess variability in output and perplexity as a measure of certainty.

\subsection{Interaction:} Although quality of interaction with a LLM is largely subjective across users and tasks, objective metrics have emerged to aid in comparative assessments. Applied effectively to presentation, metrics of readability~\cite{marulli2024understanding}, fluency~\cite{ahmed2025text}, and correspondence allow for benchmarking of output against reference standards (e.g., Kincaid reading-level). Additionally, several metrics have been developed specifically for LLM behaviors, including attributable to identifiable sources (AIS) for source linking~\cite{rashkin2023measuring}. Outside established instruments, LLM-as-a-judge approaches continue to be valued in open-ended assessment, including performance of chain-of-thought reasoning~\cite{zhang2024chain}. Finally, while assessment of LLM integrations are often descriptive, a growing set of metrics for AI-agents can quantify models’ access and use of external resources~\cite{guo2024stabletoolbench} alongside interface usability~\cite{vlachogianni2022perceived}.

\subsection{Operations:} Assessment of operational utility leverages metrics that quantify the economic and computational feasibility of adopting and maintaining LLM systems. Cost-oriented benchmarks measure tradeoffs between expenditure and model effectiveness (e.g., CEBench~\cite{sun2024cebench}), capturing whether gains in performance justify operational expenses under realistic economic constraints. Benchmarks such as Time to First Token (TTFT) and Inter-token Latency (ITL)~\cite{nvidia_nim_metrics_2025}, assess a model’s capacity to deliver timely and scalable outputs across varying conditions. Supporting metrics like Smooth Goodput~\cite{wang2024revisiting} can be used to evaluate a LLM's ability to handle concurrent requests without reducing performance. Together, these metrics anchor operations to practice, revealing how complexity and scale drive deployment viability.

\subsection{Governance:} A growing landscape of governance-oriented evaluation tools have emerged to assess compliance and security in deployed LLM systems. Recent work has introduced JailBreakV, which evaluates a model’s resistance to jailbreaking and adherence to safety constraints established by the EU AI Act~\cite{luo2024jailbreakv}. Meanwhile, measures have been established to assess  presence and severity of harmful content (e.g., HarmMetric~\cite{yang2025harmmetric}), especially when safeguards are bypassed. Security evaluations, by contrast, increasingly focus on preventative measures and detection-based defenses, with tools such as PII-Scope~\cite{kanth2024pii} and PersistBench~\cite{pulipaka2026persistbench} categorizing model risks and identifying sensitive user prompts. Collectively, these approaches enable comparative assessment of how LLMs enforce constraints under realistic and even adversarial conditions.

\end{document}